\documentclass[journal]{IEEEtran}
\ifCLASSINFOpdf
\else
\fi
\usepackage{graphicx}
\usepackage{hyperref}
\usepackage[table]{xcolor}

\begin{document}
%
\title{Codenames as a Benchmark for Large Language Models}
%
%
%

\author{Matthew Stephenson,
        Matthew Sidji,
        and Benoît Ronval
        
\thanks{Matthew Stephenson is with the College of Science and Engineering, Flinders University, Adelaide,
SA 5042, Australia (e-mail: matthew.stephenson@flinders.edu.au).}
\thanks{Matthew Sidji is with Faculty of Engineering and IT, University of Melbourne, Melbourne, VIC 3000, Australia (e-mail: matthew.sidji@unimelb.edu.au).}
\thanks{Benoît Ronval is with the ICTEAM, UCLouvain, Louvain-la-Neuve, Belgium (e-mail: benoit.ronval@uclouvain.be).}

}

\maketitle

\begin{abstract}
In this paper, we propose the use of the popular word-based board game Codenames as a suitable benchmark for evaluating the reasoning capabilities of Large Language Models (LLMs). Codenames presents a highly interesting challenge for achieving successful AI performance, requiring both a sophisticated understanding of language,  theory of mind, and epistemic reasoning capabilities. Prior attempts to develop agents for Codenames have largely relied on word embedding techniques, which have a limited vocabulary range and perform poorly when paired with differing approaches. LLMs have demonstrated enhanced reasoning and comprehension capabilities for language-based tasks, but can still suffer in lateral thinking challenges.
We evaluate the capabilities of several state-of-the-art LLMs, including GPT-4o, Gemini 1.5, Claude 3.5 Sonnet, and Llama 3.1, across a variety of board setups. Our results indicate that while certain LLMs perform better than others overall, different models exhibit varying emergent behaviours during gameplay and excel at specific roles. We also evaluate the performance of different combinations of LLMs when playing cooperatively together, demonstrating that LLM agents are more generalisable to a wider range of teammates than prior techniques.

\end{abstract}

\begin{IEEEkeywords}
Codenames, Large Language Models, Game Playing Agents, AI Benchmarks.
\end{IEEEkeywords}

%
\IEEEpeerreviewmaketitle

\section{Introduction}
\IEEEPARstart{I}{n} recent years, Large Language Models (LLMs) have seen rapid advancement, adoption and experimentation across a wide range of research fields \cite{zhao2024surveylargelanguagemodels}. Games are no exception to this, with many researchers attempting to utilise this new technology for novel game playing and content creation applications \cite{Yang_2024,10.1145/3582437.3587211}. The emergent reasoning capabilities of LLMs, as demonstrated across various cognitive and symbolic tasks, has led to further investigations into the abilities of LLMs to not only enhance the in-game experience, but also to play games autonomously \cite{10645579,jeurissen2024playingnethackllmspotential,ciolino2020transformernaturallanguagemodeling,kim2023generativeaimafialikegame,10.1145/3649921.3650013}. However, the complex spatial reasoning and strategic planning aspects of most traditional board and video games are known to be particularly challenging for LLMs \cite{costarelli2024gamebenchevaluatingstrategicreasoning}. 


In contrast to pure strategy games such as Chess or Go that have been used as AI benchmarks \cite{silverGeneralReinforcementLearning2018, silverMasteringGameGo2017b}, the language centred nature of LLMs make them much more suited to games that utilise natural language as part of the core gameplay. Many modern multi-player games permit natural language conversations between players as a means to discuss strategies or form alliances, but this aspect is often not considered by traditional AI agents \cite{gainaTAGPandemicCompetition2022}. LLMs offer a new approach to developing agents for language-based games, as well as games where players can benefit from being able to communicate with each other. 

To explore the potential of LLMs for language-based games, we have selected the board game Codenames (Czech Games Edition, 2015) as our benchmark \cite{chvatilCodenames2015}. Codenames is a popular team-based party game that requires cooperation, natural language understanding, and epistemic reasoning abilities to play effectively. Players work in asymmetric two-person teams, where one player (codemaster) provides a single word clue that attempts to link a variety of other words together. The second player (guesser) must then select from a board of several possible words, those which they feel are most associated with the provided clue. The full rules for playing Codenames are described in Section II, for any readers who are unfamiliar with this game. 

A simplified version of Codenames was previously utilised for a short-lived AI competition in 2019, that focused on developing more traditional natural language processing techniques (such as semantic word associations) \cite{original_competition}. While these approaches performed well when playing with another agent using the same word association strategy, they perform significantly worse when paired with a teammate utilising an alternative technique \cite{kimCooperationCodenamesUnderstanding2019}. LLMs may provide a solution to this problem, demonstrating emergent natural-language and theory of mind reasoning capabilities across a wide range of prior domains \cite{kosinskiTheoryMindMay2023}.

In this paper, we present an updated version of the Codenames AI framework that replicates the full rules of the original board game. We then benchmark the performance of several state-of-the-art LLMs, as well as the more traditional word-vector approaches, to play Codenames alongside a variety of teammates and opponents. This evaluation explores whether LLMs are inherently able to understand the rules of the game provided to them, along with their ability to provide meaningful and generalisable clues that work for a range of potential teammates. Our results demonstrate that, while current LLMs are not able to outperform more traditional NLP agents when cooperating with an identical technique, their performance is significantly less hindered when paired with other agents. It is also apparent that each of the LLMs tested have a different emergent playstyle, with some playing more cautiously or risky than others, leading to interesting result combinations when paired together. We also provide preliminary results for OpenAI's recent o1-preview model, which demonstrated a significant performance improvement over other LLMs.

The remainder of this paper is organised as follows: Section II details the complete rules for Codenames, along with the differences between the previous and updated version of the Codenames AI framework. Section III covers related work, including alternative LLM benchmarks, the reasoning skills required to play Codenames effectively, and the previous AI approaches for this game. Section IV provides details on the experimental analysis used to evaluate the game-playing abilities of several state-of-the-art LLMs for different versions of Codenames. Section V presents the results of these experiments for both LLMs and word-vector agents. Section VI discusses the implications of our quantitative results, along with qualitative observations and potential limitations. Section VII describes pathways for future research, with Section VIII providing the final conclusion.

\section{Codenames AI Framework}

Before discussing related work, we will first explain our updated Codenames AI framework.
One of the contributions of this research is that the previous Codenames AI framework has been extended to replicate the full game rules, along with support for LLM controlled agents. This section describes the rules of Codenames for readers who are unfamiliar with it, along with the necessary changes made to the Codenames AI framework to support the complete game ruleset. The updated framework code is available, alongside our presented results, at the following repository.\footnote{\url{https://github.com/stepmat/Codenames_GPT/tree/ToG_2025}}

\subsection{Codenames Rules}

The following four sub-sections describe the complete rules for Codenames. For additional clarification, Figure \ref{fig:Board_seed_1} shows an example board setup from the codemaster's perspective. The guesser is presented the same board but with the identities (i.e., colours) of the words hidden.

\subsubsection{Overview}
Codenames is a word-based game of language understanding and communication.
Players are split into two teams (red and blue), with each team consisting of a Codemaster and Guesser. The red team always goes first.
\subsubsection{Setup}
At the start of the game, the board consists of 25 English words.
The Codemaster on each team has access to a hidden map that tells them the identity of all of the words (Red, Blue, Civilian or Assassin). 
A standard map in Codenames has 9 red words, 8 blue words, 7 civilian words and 1 assassin word. 
The Guessers on each team do not have access to this map, and so do not know the identity of any words.
Players need to work as a team to select all their words in as few turns as possible, while minimising the number of incorrect guesses.
\subsubsection{Turns}
At the start of each team's turn, the Codemaster supplies a clue and a number (the number of words related to that clue).
The clue must:
\begin{itemize}
    \item Be semantically related to the words the Codemaster wants their Guesser to guess.
    \item Be a single English word.
    \item Not derive, or be derived from, one of the words on the board.
\end{itemize}
The clue number must be greater than or equal to zero.
The Guesser then selects from the remaining words on the board, based on which word is most associated with the Codemaster's clue.
The identity of the selected word is then revealed to all players.
If the Guesser selected a word that is their team's colour, then they may get to select another word.
The Guesser must always make at least one guess each turn, and can guess up to one word more than the number provided in the Codemaster's clue.
The only exception to this is if the Codemaster's clue number is zero, then there is no limit on the maximum number of guesses.
If a Guesser selects a word that is not their team's colour, their turn ends.
The Guesser can choose to stop selecting words (ending their turn) any time after the first guess.
\subsubsection{Ending}
Play proceeds, passing back and forth, until one of two outcomes is achieved:
\begin{itemize}
    \item All of the words of a team's colour have been selected (this team wins).
    \item A team's guesser selects the assassin word (this team loses).
\end{itemize}

\begin{figure}
\begin{center}
\scriptsize
\begin{tabular}{|p{1.3cm}|p{1.3cm}|p{1.43cm}|p{1.3cm}|p{1.35cm}|} 
 \hline
 \textcolor{darkgray}{SINK} & \textcolor{darkgray}{POOL} & \textcolor{violet}{KNIFE} & \textcolor{cyan}{ALPS} & \textcolor{cyan}{CONTRACT} \\ 
 \hline
 \textcolor{cyan}{CAR} & \textcolor{red}{PLATE} & \textcolor{red}{TRUNK} & \textcolor{cyan}{WORM} & \textcolor{red}{RULER} \\ 
 \hline
 \textcolor{cyan}{BELT} & \textcolor{red}{CHINA} & \textcolor{cyan}{PARACHUTE} & \textcolor{cyan}{FIGHTER} & \textcolor{red}{SPELL} \\ 
 \hline
 \textcolor{red}{PRESS} & \textcolor{red}{LION} & \textcolor{darkgray}{JAM} & \textcolor{darkgray}{MAPLE} & \textcolor{red}{BEACH} \\ 
 \hline
 \textcolor{darkgray}{CHICK} & \textcolor{cyan}{MOUTH} & \textcolor{darkgray}{EMBASSY} & \textcolor{darkgray}{LEMON} & \textcolor{red}{SCHOOL} \\ 
 \hline
\end{tabular}
\end{center}
\caption{Codenames example board setup (seed = 0). Words associated with each team are shown in red or blue, civilian words are shown in grey, and the assassin word is shown in purple.} 
\label{fig:Board_seed_1} 
\end{figure}

\subsection{Differences from Previous Framework}

The previous Codenames AI framework provided a simplified version of the above rules, which differs from the full game in two key aspects:

Firstly, the previous framework provided a single team cooperative version of the game, where only the red team gives clues and makes guesses. The red team is then scored at the end of the game based on the number of turns needed to select all red words (i.e., a lower score is better) with a loss resulting in a maximum score of 25 points. This rule simplification makes it easier to evaluate a single team of agents without needing to worry about the opposing team, but also removes some of the game's deductive and strategic reasoning elements. Blue and civilian words function almost identically in this single team version, with the only difference being that the red team loses if they somehow select all blue words (which essentially never happens). This lowers the potential impact that incorrect word selections have on a team's chance of winning, and reduces the game's overall strategic depth. Team's being given the maximum score of 25 points if they lose also means that selecting the assassin word can have a huge impact on a team's average overall performance, and thus incentives slow and cautious strategies. Our revised framework provides the option for both the single team (previous framework) and two teams (full rules) versions of Codenames.

Secondly, the previous framework did not permit the guesser to deviate from the number of guesses specified in the codemaster's clue. The original Codenames rules state that, assuming that an incorrect word is not selected, the guesser can select any number of words between one and one more than the number of guesses specified by the codemaster (e.g., if the codemaster's clue specifies the number three, then the guesser can select a minimum of one and a maximum of four words). The codemaster is also able to provide a clue number of zero, which allows the guesser to make as many guesses as they like (although this almost never happens in regular play). Our updated framework allows the guesser to stop guessing after each word selection, up to one more than the clue number provided by the codemaster. This provides another interesting strategic choice for the guesser, allowing them to stop early if they are unsure what word to select next or stop late if they wish to make a risky extra guess.

\subsection{Framework Limitations}



Despite our best attempts to adhere to the full Codenames rules described above, there are still some remaining ambiguities that need to be addressed. Firstly, the rules of Codenames state the the provided clue cannot derive, or be derived from, one of the words on the board. What exactly counts as a derivative word is a subjective decision, and thus cannot be easily defined in our framework. In the official rulebook for Codenames, this invalid clue rule extends to include compound words or non-English words, using the clue number as an additional hint, and providing clues that do not relate to the meaning of words. These requirements are hard to define objectively, and have thus been excluded from our framework. The only restriction we enforce is that clues cannot contain or be contained within any of the words still available on the board (i.e., no substrings allowed). This does mean that agents could effectively ``cheat'' by subverting this rule (such as deliberately misspelling words) although this is unlikely to occur without deliberate human influence.

Another rule from the original Codenames game relates to the penalty for invalid clues. In the official rulebook, if the codemaster gives an invalid clue their turn ends immediately and the other team's codemaster gets to identify one of their team's words for free. However, the imperfect nature of LLMs means that they have an increased risk of providing a clue in an invalid format (such as providing additional text in their response). We therefore decided to relax this rule, and instead simply ask the codemaster to try again if their original response was invalid. However, If the codemaster fails to provide a valid clue 10 times in a row, then a default empty string is chosen as the clue with the clue number being set to 1. Likewise, if the guesser fails to guess a valid word on the board 10 times in a row, then one of the remaining words is chosen at random. During our experiments, no codemaster ever failed to give a valid clue within this defined limit. However, the guesser agent would very occasionally fall into a repeated loop of providing a response that wasn't on the board and would eventually have a word selected at random (although this happened only a handful of times across all experiments). While this modification provides a more lenient version of the game's rules, it is important to highlight that the creation of LLMs that can accurately adhere to their provided prompts is a strict requirement of reliable AI. Future versions of this benchmark may choose to remove this modification in order to emphasise this rule following aspect of LLM evaluation.

\section{Related Work}



In this section, we discuss alternative benchmarks that have previously been used to evaluate the performance of LLMs, and why we feel that Codenames is a novel and suitable benchmark alternative. We also present some of the previous approaches to developing AI approaches for Codenames.

\subsection{Benchmarks for LLMs} 

With the ever growing number of LLMs being released every month, being able to empirically evaluate their performance has become an increasingly critical challenge. To understand how good these models are across different domains and applications, researchers often rely on multiple benchmarks that focus on different LLM capabilities or tasks.

One of the most popular benchmarks for evaluating the general knowledge and language understanding of LLMs is the Massive Multitask Language Understanding (MMLU) test \cite{hendrycks2020measuring}. This benchmark contains numerous multiple-choice questions grouped into 57 individual topics, that LLMs are required to interpret and answer correctly. Other benchmarks such as HellaSwag \cite{zellers2019hellaswag} or BIG-Bench Hard (BBH) \cite{suzgun2022challenging} also focus on language comprehension, but additionally assess general and common-sense reasoning abilities \cite{huang2022towards}. For HellaSwag, given the start of the sentence, the LLM must select the logical completion among different available choices. BBH considers 23 task groups, each consisting of several examples, that have been identified as especially challenging LLMs.

As a subcategory of reasoning, LLMs can also be evaluated on their strategic capabilities \cite{zhang2024llm}. This ranges from their playing skill for different types of games (conversational, board, card, or electronic games) to societal and economic simulations. One of the key aspects for this type of evaluation is the presence of at least one other agent that can influence the environment, which thus affects the decisions made by the LLM.
Beyond evaluating general reasoning or text generation, benchmarks have also been proposed to assess the capabilities of LLMs for specific tasks or scenarios. TyDi QA \cite{clark2020tydi} is another benchmark consisting of questions and expected answers but presented in a variety of world languages, intended to evaluate the multilingual abilities of LLMs. Another example of a precise domain benchmark is the MATH test set \cite{hendrycks2021measuring}, which (as the name would imply) assesses LLM performance on different mathematical problems.



This section has covered only a handful of the many LLM benchmarks that currently exist, however it is apparent that few of them evaluate more abstract language understanding and reasoning capabilities of LLMs outside of knowledge assessment. We propose that the use of Codenames as an LLM benchmark allows for the simultaneous assessment of language understanding, strategic reasoning, and theory of mind capabilities.


\subsection{Reasoning Skills in Codenames} 






In this section, we highlight some of the the important reasoning skills required to play Codenames effectively, and why this makes it a novel and effective benchmark for evaluating LLMs. 



The first type of reasoning is that of language or word understanding. We further distinguish this into inductive reasoning for the codemaster and deductive reasoning for the guesser. We designate the generation of a clue by the codemaster as inductive reasoning, as it requires forming a general concept (the clue) from specific provided examples (the words on the board). This can also be seen as an optimisation problem, where the codemaster has to find a clue related to the largest number of their team's words, while also avoiding associations with any other non-team words. In contrast, the guesser has to demonstrate deductive reasoning to identify specific words on the board based on their association with the provided clue. 
Demonstrating robust inductive and deductive reasoning remains a challenging task for LLMs, although recent works involving prompt engineering are encouraging and show the need for benchmarks that evaluate this capability \cite{sapNeuralTheoryofmindLimits2022, ullmanLargeLanguageModels2023a, sidjiPromptEngineeringChatGPT2024}.

The second type of reasoning present in Codenames is strategic reasoning. The codemaster may decide to give a more risky or cautious clue to the guesser, commonly exemplified as the number associated with the clue word. A higher clue number would allow the guesser to select more words, but can also increase the risk of them selecting one of the words for the other team (or even the assassin word). The guesser may also make strategic decisions, by deciding whether to stop guessing early when unsure of the next word, stopping at the number specified by the codemaster, or making an extra guess.
The decision whether to play risky or cautious should ideally depend on the current state of the game, as a team that is further ahead than the opposing team can afford to play safer, while a team that is behind or close to losing make choose to go for a ``Hail Mary'' final attempt. We should expect intelligent players to take all these factors into account when making their decisions and adapt to changes in relative team performance throughout the game.
Recent research has indicated that LLMs may possess the capability for generalisable strategic reasoning in simple test cases, indicating their potential suitability for Codenames \cite{Gandhi2023StrategicRW}. 

Lastly, Codenames also requires cooperative and epistemic reasoning. Both the codemaster and the guesser have to play with each other in order to win, despite not being able to communicate outside of their defined actions. These Partial Information, Restricted Communication, Cooperative (PIRCC) games have recently garnered interest in the AI literature due to the complexity of creating capable AI players, and humans' current superior performance to many AI \cite{sidjiHumanAICollaborationCooperative2024}. To perform in these settings effectively, each agent needs to internalise how their teammate, and to a lesser extent the opposing team, may interpret their actions (i.e., what is their current mental model). To perform better at the game, this reasoning needs to be applied when inducing suitable clues (codemaster) and deducing words on the board (guesser). This reasoning also allows players to adapt their strategy and playstyle based on the performance and actions of their teammate.
For example, a player may give a clue about medieval history because their teammate is knowledgeable on the subject, or a highly skilled codemaster may need to reduce their clue complexity when paired with a novice guesser.
LLMs may be able to use this type of reasoning to play Codenames better than other more traditional agents, having previously demonstrated some theory of mind capabilities \cite{kosinskiTheoryMindMay2023, streetLLMsAchieveAdult2024}, although there remains debate over whether this reflects true possession of a theory of mind \cite{sapNeuralTheoryofmindLimits2022, ullmanLargeLanguageModels2023a}.


Based on the multiple interconnected reasoning capabilities involved in playing Codenames, we believe that this game provides a complex and nuanced task that assesses multiple facets of cognitive intelligence. 



\subsection{Codenames AI}

There is growing interest in the game Codenames from AI research due to its demand for multi-modal language understanding, asymmetric cooperation, theory of mind, and epistemic reasoning \cite{kimCooperationCodenamesUnderstanding2019}. Introduced by Kim et al. (2019) the first Codenames AI competition employed word embedding techniques such as word2vec and GloVe models. These models achieved 100\% accuracy when paired with themselves as teammates, but saw a drop in performance when working with teammates using different models. As an extension of this work Jaramillo et al. (2020) utilised term frequency - inverse document frequency (TF-IDF), Naive-Bayes, and the GPT-2 Transformer models. They report that the transformer model achieved the same or better accuracy compared to agents developed by Kim et al. (2019). When tested with human participants, the transformer model was preferred over other agents \cite{Jaramillo_Charity_Canaan_Togelius_2020}.

Koyyalagunta et al. (2021) developed multiple methods to improve Codenames AI performance when paired with humans. Their aims was to create agents that produce more human-interpretable clues, as previous agents would often give clues that were nonsensical to human guessers but would be correctly guessed when paired with another word embedding agent \cite{koyyalaguntaPlayingCodenamesLanguage2021}. Other researchers have focused on agents with the ability to adapt in real time to their teammates. Archibald et al. (2024) created the Adaptive Codenames Ensemble (ACE) which changes the Codenames agent it produces clues with based on the teammates guesses \cite{archibaldAdaptingTeammatesCooperative2024}. Archibald et al. (2024) also proposed a Noisy Communication Model (NCM) which deliberately adds ``noise'' to a clue in order to increase it's generalisability to unknown teammates \cite{10645589}. 

Prompting techniques for LLMs have also been used to improve Codenames AI performance. Ozturkler et al. (2023) use ThinkSum, a prompting technique used to promote deductive reasoning, to improve Codenames guesser agents. They showed a 20\% improvement in score compared to few-shot prompting for guesser agents \cite{ozturklerThinkSumProbabilisticReasoning2023}.
Most recently, Sidji et al. (2024) compared the performance of various prompt engineering approaches on Codenames using OpenAI's GPT-4-1106 as the base LLM. This included prompting techniques such as Chain of Thought \cite{kojimaLargeLanguageModels2022a}, Self Refine \cite{madaanSelfRefineIterativeRefinement2023}, and Solo Performance \cite{wangUnleashingEmergentCognitive2024, sidjiPromptEngineeringChatGPT2024}. While these prompting techniques had a measurable impact on the agent's playstyle, typically resulting in more risky or cautious clues, none of them were able to produce a significant improvement to overall performance. Rather than exploring the effectiveness of prompt engineering techniques, our presented research instead aims to explore the inherent abilities of different state-of-the-art LLMs to play Codenames effectively when paired with or against other alternative models.

\section{Experiments}

This section describes the selected AI agents, Codenames game versions, and evaluation process carried out using our updated Codenames AI framework. The primary purpose of these experiments was to determine the current performance of state-of-the-art LLMs for playing Codenames, compared to traditional word-vector approaches.

\subsection{Game Playing Agents}

\subsubsection{LLM Agents}

The following LLM families and versions were initially selected for evaluation:
\begin{itemize}
  \item GPT (OpenAI)
  \begin{itemize}
    \item \textbf{o1-preview (2024-09-12)}
    \item \textbf{o1-mini (2024-09-12)}
    \item \textbf{o3-mini (2025-01-31)}
    \item \textbf{GPT-4o (2024-08-06)}
    \item GPT-3.5-turbo (0125)
  \end{itemize}
  \item Gemini (Google Deepmind)
  \begin{itemize}
    \item \textbf{Gemini-1.5 (Pro 002)}
  \end{itemize}
  \item Claude (Anthropic)
  \begin{itemize}
    \item \textbf{Sonnet-3.5 (2024-10-22)}
    \item Haiku-3.5 (2024-10-22)
  \end{itemize}
  \item DeepSeek
  \begin{itemize}
    \item \textbf{DeepSeek-R1 (2025-01-20)}
    \item \textbf{DeepSeek-V3 (2024-12-26)}
  \end{itemize}
  \item Llama (Meta)
  \begin{itemize}
    \item \textbf{Llama-3.1 (70B-Instruct)}
    \item Llama 3.2 (3B-Instruct)
  \end{itemize}
  \item Phi (Microsoft)
  \begin{itemize}
    \item Phi-3-medium (128k-Instruct)
  \end{itemize}
  \item Mistral AI (Mistral)
  \begin{itemize}
    \item Mistral-0.3 (7B-Instruct)
    \item Mixtral-0.1 (8x7B-Instruct)
  \end{itemize}
\end{itemize}

However, it became apparent during preliminary testing that many of the smaller models were not able to correctly adhere to the game's specified rules and output format, resulting in repeated invalid clues or guesses that frequently devolved into pure random play. As such, only the nine LLMs highlighted in bold (o1-preview, o1-mini, o3-mini, GPT-4o, Gemini-1.5, Sonnet-3.5, DeepSeek-R1, DeepSeek-V3 and Llama-3.1) were able to consistently follow the game's rules. Scores for the other LLMs were significantly worse because of this, and we thus chose to only present our findings and analysis for these nine higher performing LLMs.

\subsubsection{Word-Vector Agents}

In addition to the LLM agents mentioned above, we also evaluated three word-vector agents supplied with the previous Codenames AI framework: 
\begin{itemize}
  \item Word2Vec (threshold = 0.7).
  \item GloVe (300d)
  \item Combined (300d, threshold = 0.7)
\end{itemize}

These agents utilise word embedding approaches based on learned vector semantics from a provided training corpus \cite{kimCooperationCodenamesUnderstanding2019}. \textbf{Word2Vec} is based on a 300 dimensional pre-trained skip-Gram model trained on the Google News corpus \cite{mikolov2013efficientestimationwordrepresentations}. \textbf{GloVe} (Global Vectors for Word Representation) is an alternative approach that additionally considers the co-occurrence of words within a defined context \cite{pennington-etal-2014-glove}. \textbf{Combined} utilises both approaches by concatenating the Word2Vec and GloVe vectors together \cite{rücklé2018concatenatedpowermeanword}.

One crucial limitation of these word-vector agents is that, due to the fact that their approaches utilise a pre-defined corpus of words for determining suitable clues / guesses, they are unable to interpret any words which are not present in their training set. Preliminary testing found that these agents will often produce errors when paired with LLM teammates, which have a much wider ranging vocabulary from which to select clues. Because of this, we were unable to produce reliable performance results for games with both LLM and Word-Vector agents on the same team, and have instead chosen to evaluate each agent group separately.

\subsection{Game Versions}

Two different versions of Codenames were utilised for our agent evaluations, based on the previous and updated versions of the Codenames AI framework.

\subsubsection{Single Team (cooperative)}
Played using the same scoring system as the previous Codenames AI framework, where a single team (red codemaster/guesser) attempts to identify all red words in as few turns as possible. Teams are awarded a score at the end of the game based on the number of turns taken (lower score is better). The only exception to this is if the guesser selects all blue words or the assassin word, which results in a maximum score of 25 points. 

\subsubsection{Two Teams (competitive / cooperative)}
Played using the full set of rules from the original Codenames game, where two teams (red codemaster/guesser and blue codemaster/guesser) attempt to identify all words of their team's colour first. Selecting the assassin word results in an immediate win for the other team. Guessers can also inadvertently help the opposing team win if they accidentally select any words of their colour. Rather than using a scoring system, this version measures success in terms of overall win-rate.

\subsection{LLM Agent Prompts}

When an LLM agent is initialised within the Codenames AI framework it is first provided with an input prompt describing the game's rules, using the same format and wording shown in section II.A, along with its team's colour and whether it is playing as the codemaster or guesser (e.g., ``You are playing the game Codenames as the Red Guesser''). When required to give a response during the game, each agent is prompted as follows:

\subsubsection{Codemaster}
The only situation where the codemaster needs to provide a response is when giving a clue (word and number) based on the current board state. The codemaster is provided with the remaining words on the board ordered by their identity (red, blue, civilian and assassin) along with an instruction to ``provide a single word clue and number for the guesser in the following format (\textquotesingle pebble\textquotesingle,2)''. The codemaster is also reminded that ``the clue cannot be derived from or derive one of the words on the board'', and informed that they are to ``stick to this format exactly and provide no additional text''. If the provided clue format is invalid or violates the ``no derived words'' rule, the codemaster is informed of this and asked to respond again.

\subsubsection{Guesser}
The guesser has two possible situations where they are required to give a response: selecting a word on the board that matches the provided clue, and deciding whether to continue guessing after a correct guess has been made. 
When required to select a word for their guess, the guesser is provided with all the remaining words on the board as well as the codemaster's clue (e.g., ``(pebble,2)''). The guesser is instructed to ``select one of the remaining words that is most associated with this clue'', and informed that they ``must select one of the remaining words and provide no additional text''.
When deciding whether to continue guessing, the guesser is once again informed of the remaining words on the board and the codemaster's clue. The guesser is also told how many words it has already picked this turn, and is then asked ``Would you like to keep guessing? Answer only \textquotesingle yes\textquotesingle or \textquotesingle no\textquotesingle''. If an invliad response is provided in either of these cases, the guesser is informed of this and asked to respond again.

\subsection{Evaluation Procedure}

Based on the agent and game versions described above, the following evaluation experiments were conducted:

\subsubsection{Single Team}
This experiment compared the performance of different LLM and word-vector agent combinations, acting as both the codemaster and guesser, for the single team version of Codenames.
Due to the previously mentioned issues with word-vector agents failing to understand clues provided by LLM agents, these two groups of agents were evaluated separately.
In addition, due to the high usage costs currently required to run OpenAI's recent o1/o3 models, the o1-preview, o1-mini and o3-mini LLMs were only evaluated when paired with themselves. Likewise, the DeepSeek models were also only evaluated when paired with themselves, although this was due to imposed usage limits at the time of writing.

\subsubsection{Two Teams}
This experiment evaluated how different teams of the same agent (i.e., identical models for the codemaster and guesser) performed on the new two teams version of Codenames (i.e., the full Codenames rules). While it would have been beneficial to compare how teams of different agents also performed, this would have led to an exponential increase in the number of agent combinations and associated costs. Similar to the single team version, LLM and word-vector agent groups were evaluated separately. We also chose to exclude the o1 models from this experiment due to the aforementioned high usage costs.

Each of the agent and game version combinations mentioned above were evaluated over 100 trials (using random seeds 0-99 inclusive). 
All code utilised for this experiment, along with full quantitative results, input prompts and output responses, is available in the provided public code repository. All experiments were conducted on an AMD Ryzen Threadripper PRO 5955WX, with an RTX A6000 and 256GB of RAM (although the specific hardware utilised should not have an impact on the produced results). Due to the unpredictable nature of the LLMs used, exact replication of our results is likely not possible even if the same parameters are applied. However, we have found that the general trends in our results are consistent over multiple experiment runs. 

Regarding the costs associated with replicating these experiments, while the Llama models are open-source, the GPT, Gemini and Claude models are currently only available using a paid service offered by each provider. As of the time of writing, the costs incurred as a result of these experiments total approximately \$538 USD for GPT (OpenAI), \$141 USD for Gemini (Google Deepmind), \$177 USD for Claude (Anthropic) and \$8.33 USD for DeepSeek. Please note that this does not include the additional subscription fee required to access the OpenAI API. The bulk of these costs was attributed to the new o1-preview model provided by OpenAI, which cost \$288 USD alone for running 100 trials of the single team game version. While costs for using these models are expected to reduce over time, this can still be a prohibitively expensive requirement for any students or self-funded researchers wishing to experiment with these same models.

\section{Results}

This section provides results for the agent evaluation experiments described above.

\subsection{Single Team Version}

Summative results for experiments utilising the single team version of Codenames are shown in Table \ref{results_table1}. Definitions for each column are provided as follows:
\begin{itemize}
    \item \textbf{Model Pair:} Defines the two models being applied, with the former specifying the codemaster and the latter specifying the guesser (i.e., codemaster - guesser). Results for each row were calculated from 100 individual trials.
    \item \textbf{Mean:} Mean score. Note, a team's score is equal to the number of turns needed to identify all red words (i.e., a lower value is better), with the exception that selecting the assassin word results in a score of 25 points.
    \item \textbf{Median:} Median score.
    \item \textbf{Min:} Minimum score.
    \item \textbf{Std Dev:} Standard deviation in the score.
    \item \textbf{Loss:} Percentage of games that ended in a loss. Note, for our single team results all losses were caused by selecting the assassin word (rather than selecting all of the blue words). 
    \item \textbf{Mean (without loss):} Mean score across all games that did not end in a loss (i.e., games where all red words were identified).
    \item \textbf{Blue avg:} Average number of blue words selected in each game (standard deviation provided in brackets).
    \item \textbf{Civilian avg:} Average number of civilian words selected in each game (standard deviation provided in brackets).
    \item \textbf{Clues avg:} Average clue number given by the codemaster each turn (standard deviation provided in brackets).
    \item \textbf{Guesses avg:} Average number of guesses made by the guesser each turn (standard deviation provided in brackets).
    \item \textbf{Stop Early:} Percentage of turns where the guesser chose to voluntarily stop guessing before reaching the clue number provided by the codemaster.
    \item \textbf{Stop Late:} Percentage of turns where the guesser chose to guess one word more than the clue number provided by the codemaster.
\end{itemize}

\begin{table*}
\caption{Agent Results for Single Team Codenames Version} 
\label{results_table1} 
\begin{center}
\scriptsize
\begin{tabular}{|p{3.00cm}|p{0.6cm}|p{0.8cm}|p{0.45cm}|p{0.5cm}|p{0.5cm}|p{0.9cm}|p{1.1cm}|p{1.1cm}|p{1.1cm}|p{1.1cm}|p{0.6cm}|p{0.6cm}|} 
 \hline
 \textbf{Model Pair \newline (codemaster - guesser)} & \textbf{Mean} & \textbf{Median} & \textbf{Min} & \textbf{Std Dev} & \textbf{Loss} & \textbf{Mean (without loss)} & \textbf{Blue avg(stdev)} & \textbf{Civilian avg(stdev)} & \textbf{Clues avg(stdev)} & \textbf{Guesses avg(stdev)} & \textbf{Stop Early} & \textbf{Stop Late}  \\ 
  \hline\hline
  o1-preview - o1-preview & 8.41 & 6 & 4 & 6.53 & 13\% & 5.93 & 0.84 (0.83) & 1.20 (1.08) & 1.95 (0.69) & 1.88 (0.70) & 11.0\% & 14.2\% \\
 \hline
 o1-mini - o1-mini & 11.10 & 7 & 4 & 7.93 & 24\% & 6.71 & 1.74 (1.19) & 1.80 (1.24) & 2.09 (0.70) & 1.91 (0.81) & 11.6\% & 16.0\% \\ 
 \hline
  o3-mini - o3-mini & 9.70 & 6 & 4 & 7.60 & 19\% & 6.11 & 1.68 (1.05) & 1.70 (1.23) & 3.40 (1.69) & 2.04 (0.95) & 17.4\% & 0.4\% \\ 
 \hline
 \hline
 DeepSeek-R1 - DeepSeek-R1 & 11.24 & 7 & 4 & 8.11 & 25\% & 6.65 & 1.76 (1.28) & 2.04 (1.18) &  2.95 (1.09) & 1.97 (0.85) & 16.0\% & 0.0\% \\
 \hline
 DeepSeek-V3 - DeepSeek-V3 & 10.59 & 8 & 5 & 6.42 & 16\% & 7.85 & 1.43 (1.09) & 1.23 (1.04) & 1.77 (0.72) & 1.50 (0.52) & 14.7\% & 0.0\% \\
 \hline
 \hline
 GPT-4o - GPT-4o & 10.58 & 8 & 6 & 6.41 & 16\% & 7.83 & 0.96 (0.97) & 1.25 (1.13) & 1.51 (0.60) & 1.45 (0.56) & 0.0\% & 0.0\% \\
 \hline
 GPT-4o - Gemini-1.5 & 9.81 & 8 & 6 & 5.21 & 10\% & 8.12 & 1.19 (1.0) & 1.01 (0.88) & 1.50 (0.60) & 1.39 (0.56) & 6.1\% & 0.5\% \\
 \hline
 GPT-4o - Sonnet-3.5 & 10.50 & 8 & 6 & 6.20 & 15\% & 7.94 & 1.26 (1.16) & 1.13 (0.86) & 1.52 (0.61) & 1.45 (0.57) & 2.0\% & 1.6\% \\
 \hline
 GPT-4o - Llama-3.1 & 11.17 & 9 & 5 & 6.43 & 17\% & 8.34 & 1.43 (1.25) & 1.18 (0.94) & 1.49 (0.60) & 1.41 (0.56) & 3.1\% & 3.8\% \\
 \hline
  \hline
 Gemini-1.5 - GPT-4o & 10.95 & 8 & 5 & 6.55 & 17\% & 8.07 & 1.42 (1.23) & 1.35 (1.14) & 1.62 (0.72) & 1.49 (0.64) & 0.3\% & 0.0\% \\ 
 \hline
 Gemini-1.5 - Gemini-1.5 & 10.38 & 8 & 5 & 6.08 & 14\% & 8.00 & 1.36 (1.06) & 1.01 (0.88) & 1.63 (0.73) & 1.46 (0.62) & 3.9\% & 0.2\% \\ 
 \hline
 Gemini-1.5 - Sonnet-3.5 & 11.18 & 8 & 5 & 6.84 & 19\% & 7.94 & 1.42 (1.08) & 1.15 (0.97) & 1.62 (0.72) & 1.51 (0.65) & 2.9\% & 2.8\% \\ 
 \hline
 Gemini-1.5 - Llama-3.1 & 11.28 & 9 & 5 & 6.46 & 17\% & 8.47 & 1.76 (1.45) & 1.51 (1.10) & 1.65 (0.71) & 1.48 (0.63) & 3.7\% & 1.8\% \\
 \hline
  \hline
 Sonnet-3.5 - GPT-4o & 12.34 & 7 & 5 & 8.39 & 30\% & 6.91 & 1.94 (1.05) & 1.87 (1.27) & 2.04 (0.59) & 1.88 (0.58) & 3.7\% & 0.0\% \\ 
 \hline
 Sonnet-3.5 - Gemini-1.5 & 10.55 & 8 & 5 & 6.89 & 18\% & 7.38 & 1.90 (1.24) & 1.74 (1.06) & 2.03 (0.61) & 1.75 (0.64) & 18.1\% & 0.0\% \\
 \hline
 Sonnet-3.5 - Sonnet-3.5 & 11.01 & 8 & 5 & 7.55 & 22\% & 7.06 & 1.97 (1.15) & 1.81 (1.00) & 2.02 (0.60) & 1.84 (0.61) & 6.9\% & 0.6\% \\ 
 \hline
 Sonnet-3.5 - Llama-3.1 & 11.31 & 8 & 5 & 7.20 & 21\% & 7.67 & 1.95 (1.34) & 1.66 (1.13) & 1.99 (0.61) & 1.66 (0.65) & 21.3\% & 0.5\% \\
 \hline
  \hline
 Llama-3.1 - GPT-4o & 10.56 & 9 & 6 & 5.51 & 12\% & 8.59 & 1.13 (0.99) & 0.95 (0.86) & 1.34 (0.53) & 1.29 (0.50) & 0.3\% & 0\% \\
 \hline
 Llama-3.1 - Gemini-1.5 & 10.54 & 9 & 6 & 5.29 & 11\% & 8.75 & 1.01 (0.94) &  1.04 (0.93) & 1.32 (0.51) & 1.29 (0.48) & 0.8\% & 0.3\% \\
 \hline
 Llama-3.1 - Sonnet-3.5 & 10.79 & 9 & 6 & 5.66 & 13\% & 8.67 & 1.10 (0.95) & 1.01 (0.99) & 1.34 (0.53) & 1.30 (0.50) & 0.6\% & 1.6\% \\
 \hline
 Llama-3.1 - Llama-3.1 & 10.18 & 9 & 6 & 4.66 & 8\% & 8.89 & 1.28 (1.10) & 1.13 (1.02) & 1.30 (0.51) & 1.30 (0.50) & 0.1\% & 5.7\% \\ 
  \hline
  \hline
 Word2Vec - Word2Vec & 6.81 & 6 & 4 & 0.97 & 0\% & 6.81 & 0.0 (0.0) & 0.0 (0.0) & 1.46 (0.64) & 1.46 (0.64) & 0.0\% & 0.0\% \\ 
 \hline
 Word2Vec - GloVe & 12.35 & 9 & 5 & 7.14 & 22\% & 8.78 & 1.68 (1.78) & 1.38 (1.53) & 1.53 (0.65) & 1.38 (0.58) & 0.0\% & 0.0\% \\  
 \hline
 Word2Vec - Combined & 10.64 & 8 & 5 & 6.72 & 16\% & 7.90 & 0.99 (1.48) & 1.05 (1.30) & 1.48 (0.64) & 1.39 (0.60) & 0.0\% & 0.0\% \\ 
 \hline
  \hline
 GloVe - Word2Vec & 12.21 & 9 & 5 & 7.52 & 24\% & 8.17 & 2.03 (1.53) & 1.45 (1.19) & 1.74 (0.76) & 1.54 (0.69) & 0.0\% & 0.0\% \\ 
 \hline
 GloVe - GloVe & 5.24 & 5 & 4 & 0.79 & 0\% & 5.24 & 0.0 (0.0) & 0.0 (0.0) & 1.72 (0.75) & 1.72 (0.75) & 0.0\% & 0.0\% \\ 
 \hline
 GloVe - Combined & 5.42 & 5 & 4 & 0.89 & 0\% & 5.42 & 0.15 (0.39) & 0.09 (0.29) & 1.71 (0.75) & 1.70 (0.75) & 0.0\% & 0.0\% \\ 
 \hline
  \hline
 Combined - Word2Vec & 11.18 & 8 & 5 & 7.37 & 21\% & 7.51 & 1.47 (1.30) & 1.04 (1.07) & 1.66 (0.74) & 1.54 (0.69) & 0.0\% & 0.0\% \\  
 \hline
 Combined - GloVe & 5.86 & 6 & 4 & 2.18 & 1\% & 5.67 & 0.09 (0.32) & 0.12 (0.36) & 1.64 (0.73) & 1.63 (0.72) & 0.0\% & 0.0\% \\ 
 \hline
 Combined - Combined & 5.53 & 6 & 4 & 0.83 & 0\% & 5.53 & 0.0 (0.0) & 0.0 (0.0) & 1.63 (0.73) & 1.63 (0.73) & 0.0\% & 0.0\% \\ 
 \hline
\end{tabular}
\end{center}
\end{table*}

\subsection{Two Teams Version}

Summative results for experiments utilising the two teams version of Codenames are shown in Table \ref{results_table2}. Definitions for each column are provided as follows:
\begin{itemize}
    \item \textbf{Model Pair:} Defines the two models being applied for each team (both codemaster and guesser), with the former specifying the red team's models and the latter specifying the blue team's models (i.e., red vs. blue). Results for each row were calculated from 100 individual trials.
    \item \textbf{Win-rate:} Percentage of games won by the red and blue teams respectively.
    \item \textbf{Assassin losses:} Percentage of games lost due to selecting the assassin word, for the red and blue teams respectively.
\end{itemize}

\begin{table}
\caption{Agent Results for Two Teams Codenames Version} 
\label{results_table2} 
\begin{center}
\begin{tabular}{|l|p{1.6cm}|p{1.8cm}|} 
 \hline
 \textbf{Model Pair (red vs. blue)} & \textbf{Win-rate (red/blue)} & \textbf{Assassin losses (red/blue)}  \\ 
 \hline\hline
 GPT-4o vs. GPT-4o & 40\% / 60\% & 13\% / 19\% \\ 
 \hline
  GPT-4o vs. Gemini-1.5 & 59\% / 41\% & 16\% / 13\% \\  
 \hline
 GPT-4o vs. Sonnet-3.5 & 52\% / 48\% & 11\% / 19\% \\ 
 \hline
 GPT-4o vs. Llama-3.1 & 53\% / 47\% & 14\% / 9\% \\ 
 \hline
  \hline
 Gemini-1.5 vs. GPT-4o & 30\% / 70\% & 18\% / 12\% \\ 
 \hline
 Gemini-1.5 vs. Gemini-1.5 & 45\% / 55\% & 17\% / 13\% \\  
 \hline
 Gemini-1.5 vs. Sonnet-3.5 & 45\% / 55\% & 13\% / 20\% \\ 
 \hline
 Gemini-1.5 vs. Llama-3.1 & 46\% / 54\% & 17\% / 13\%  \\  
 \hline
  \hline
 Sonnet-3.5 vs. GPT-4o & 50\% / 50\% & 16\% / 20\% \\ 
 \hline
 Sonnet-3.5 vs. Gemini-1.5 & 58\% / 42\% & 24\% / 13\% \\ 
 \hline
 Sonnet-3.5 vs. Sonnet-3.5 & 58\% / 42\% & 18\% / 24\% \\ 
 \hline
 Sonnet-3.5 vs. Llama-3.1 & 61\% / 39\% & 16\% / 15\% \\ 
 \hline
  \hline
 Llama-3.1 vs. GPT-4o & 45\% / 55\% & 8\% / 21\% \\ 
 \hline
 Llama-3.1 vs. Gemini-1.5 & 53\% / 47\% & 9\% / 13\%  \\ 
 \hline
 Llama-3.1 vs. Sonnet-3.5 & 44\% / 56\% & 15\% / 19\% \\ 
 \hline
 Llama-3.1 vs. Llama-3.1 & 43\% / 57\% & 10\% / 12\%  \\ 
  \hline\hline
 Word2Vec vs. Word2Vec & 49\% / 51\% & 0.0\% / 0.0\% \\ 
 \hline
  Word2Vec vs. GloVe & 28\% / 72\% & 0.0\% / 0.0\% \\  
 \hline
 Word2Vec vs. Combined & 34\% / 66\% & 0.0\% / 0.0\% \\ 
 \hline
  \hline
 GloVe vs.Word2Vec & 79\% / 21\% & 0.0\% / 0.0\% \\ 
 \hline
 GloVe vs. GloVe & 55\% / 45\% & 0.0\% / 0.0\% \\  
 \hline
 GloVe vs. Combined & 63\% / 35\% & 0.0\% / 0.0\% \\ 
 \hline
  \hline
 Combined vs. Word2Vec & 69\% / 31\% & 0.0\% / 0.0\% \\ 
 \hline
 Combined vs. GloVe & 47\% / 53\% & 0.0\% / 0.0\% \\ 
 \hline
 Combined vs. Combined & 57\% / 43\% & 0.0\% / 0.0\% \\ 
 \hline
\end{tabular}
\end{center}
\end{table}






\section{Discussion}


\subsection{Quantitative Results}

\subsubsection{Single Team Version}

Looking first at the performance of the LLMs for the single team version of Codenames, see Table \ref{results_table1}, we can see that the o1-preview model achieved the best performance when paired with a matching model (i.e., the same base LLM for both the codemaster and guesser). 
Looking at the other LLMs, the difference in performance is not as clear, with the general trend being that model pairs with a higher loss percentage typically received a higher average score. This can largely be attributed to fact that a loss results in a very high score of 25, meaning that it is often strategically better to play it safe rather than running the risk of selecting the assassin word (at least when averaging performance over multiple trials).

Looking at the other statistics for each LLM, we can make several observations about the different play styles of each model. One major differentiating factor between the models was their level of risk taking as both codemaster and guesser. The riskiness of codemasters was estimated using the \textit{Clues Avg} column. Based on the average clue number given by each codemaster, we can order the LLMs from most to least risky (o3-mini, DeepSeek-R1, o1-mini, Sonnet-3.5, o1-preview, DeepSeek-V3, Gemini-1.5, GPT-4o, and Llama-3.1). This riskiness in terms of the clue number given also heavily correlates with the percentage number of losses, with a Spearman's rank correlation coefficient of 0.821 between these two columns when considering all LLM model pairs (i.e., riskier codemasters tended to have a high percentage number of losses, when accounting for differences in the guesser model). The \textit{Median} and \textit{Min} columns also back up these findings as these values appear generally lower for the risky codemasters, indicating that when these models do win that they do so in a lower number of turns. Risky codemasters are also likely to have a higher number of Blue and Civillian words selected.
 
The riskiness of guessers was estimated using a combination of the \textit{Stop Early} and \textit{Stop Late} values for each model (averaged across the different possible codemasters). A higher \textit{Stop Early} percentage should indicate a more cautious guesser, while a higher \textit{Stop Late} percentage would indicate a more risky guesser. Note this way of measuring guesser riskiness is not an exact science, as it also appears that some models (such as GPT-4o) are much more likely to stick to the clue number provided by the codemaster. Based on the general trends of these metrics, it would appear that Gemini-1.5 and Llama-3.1 are more likely to stop early. Models also appeared to have a higher \textit{Stop Early} percentage when paired with a risky codemaster (e.g., Sonnet-3.5) and were much lower for cautious codemasters (e.g., Llama-3.1). The o1-preview, o1-mini, o3-mini, DeepSeek-R1 and DeepSeek-V3 LLMs also had very high \textit{Stop Early} percentages, although this could have been due to the fact that they were only evaluated when paired with themselves and are also risky codemasters. However, both o1-preview and o1-mini also had very high \textit{Stop Late} percentages, indicating that these models are less likely to stick to the codemaster's clue number. Sonnet-3.5 and Llama-3.1 did appear to have a slightly higher \textit{Stop Late} percentage than other LLMs, when accounting for different codemasters, but this was still fairly low by comparison. 

\begin{figure}
    \centering
    \includegraphics[width=1.0\linewidth]{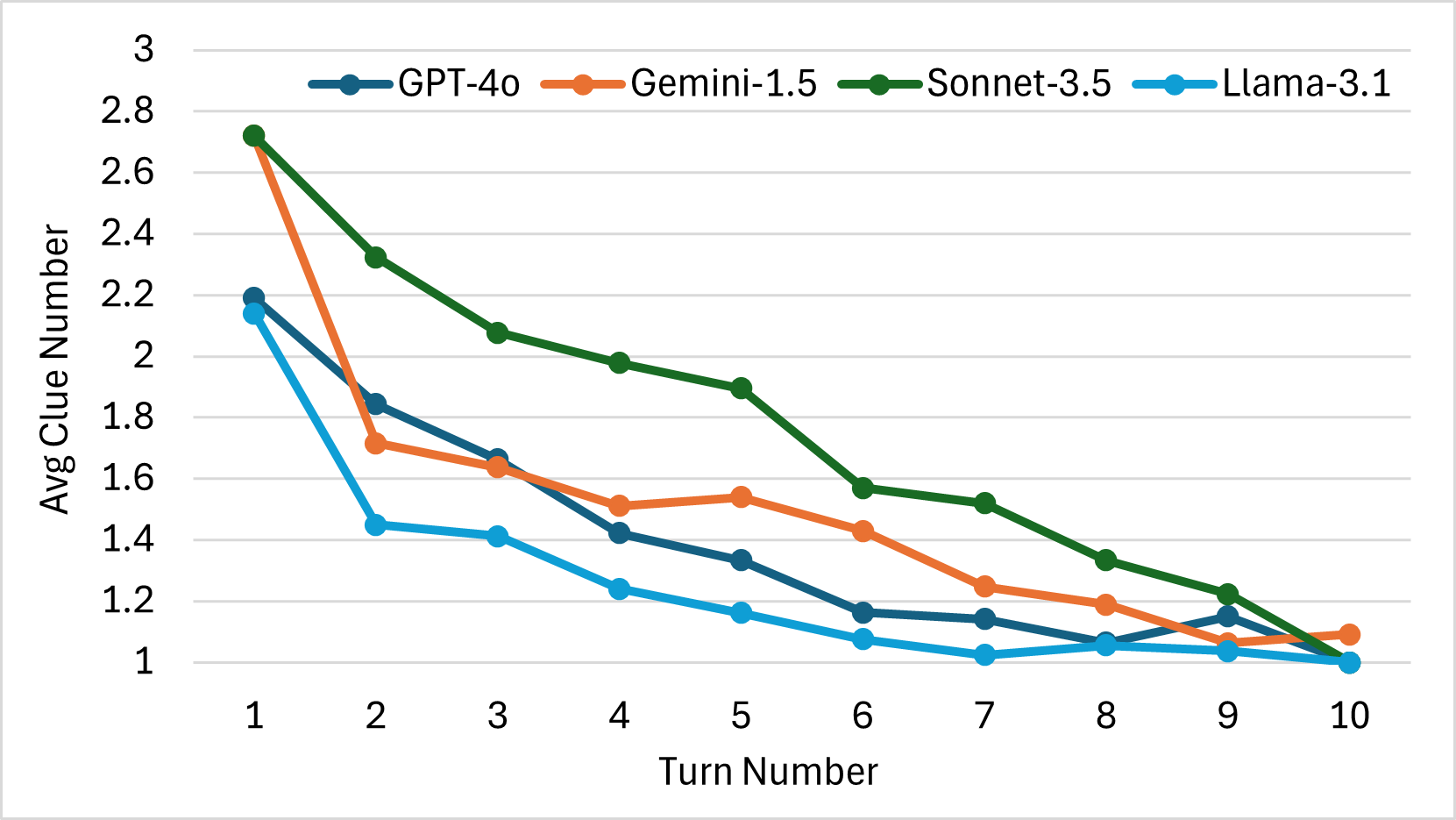}
    \caption{Average clue number provided by each codemaster model as the turn number increases (single team version).}
    \label{fig:avg_clue_numbers_codemaster}
\end{figure}

Figure \ref{fig:avg_clue_numbers_codemaster} provides further details on how the average clue number provided by each model (when acting as the codemaster) changes based on the turn number. From this we can see that our previous ordering of models from most risky to most cautious (Sonnet-3.5, Gemini-1.5, GPT-4o, and Llama-3.1) appears to hold, although the downward trends are not identical. Notably, Gemini-1.5 has an average first turn clue number of 2.72 (identical to that of Sonnet-3.5) but this value falls significantly in subsequent turns. Llama-3.1 also exhibits a similar second turn drop in clue numbers, whereas GPT-4o and Sonnet-3.5 appear to have a much more gradual decrease.


In contrast to the LLM agents, the word-vector agents consistently achieved better performance when paired with a matching partner model, resulting in a very low mean score and a 0\% loss rate. GloVe appeared to have a slightly higher average clue number than the other models, which also resulted in a slightly lower mean score. However, it is apparent that the performance of these models drops significantly when playing with a different word-vector approach (particularly when Word2Vec is paired with GloVe). This limitation does not appear to be present in the LLM agents, which performed roughly equally well when paired with other LLMs.

\subsubsection{Two Teams Version}

Looking at the result for the two teams version of Codenames, see Table \ref{results_table2}, we can see some interesting differences in the performance of certain LLMs compared to the single team version. Firstly, it would appear that riskier codemasters (e.g., sonnet-3.5) perform much better, likely due to the reduced impact that selecting an assassin word has on a team's average win-rate. Looking at the overall win-rates for both the red and blue teams, there does not appear to be a substantial difference between each side (the red team always goes first, but has nine words to identify compared to blue team's eight words). Across all trials, Sonnet-3.5 had the highest average win-rate for the red team of 56.75\% (compared to 51.00\% for GPT-4o, 41.50\% for Gemini-1.5, and 46.25\% for Llama-3.1) while GPT-4o has the highest overall win-rate for the blue team of 58.75\% (compared to 46.25\% for Gemini-1.5, 50.25\% for Sonnet-3.5, and 49.25\% for Llama-3.1). The percentage of assassin losses for each model did not appear substantially different from in the single team version, indicating little change in terms of the overall riskiness / cautiousness of each LLM's strategy. 
For the word-vector agents, Word2Vec was the worst performing overall while GloVe was the best performing (same result as for the single team version).

The observed difference in the effectiveness of different models / playstyles between the single and two teams versions of Codenames demonstrates some of the potential limitations with relying on the scoring approach of the single team version. Most notably, selecting the assassin word has a huge impact on a team's overall mean score for the single team version, which results in a game that encourages cautious play. The two team version of Codenames instead appears to incentivise the opposite, with Sonnet-3.5 going from the worst performing LLM for the single team version (in terms of mean score across all games played) to the best performing LLM for the two teams version (based on average win-rate across all games played). This would seem to imply that playing slightly riskier than your opponent is a smart strategy for this version, particularly when paired against overly cautious teams. 



\subsection{Qualitative Observations}
This section will discuss anecdotal findings regarding model behaviour that was observed throughout the games played, including several instances where it appeared the LLMs did not fully internalise the game's rules. 
We propose some possible explanations for these behaviours, although the black box nature of LLMs makes it difficult to verify the true cause.

\subsubsection{Differences between word-vector and LLM approaches}
\textbf{LLMs can give fictional word clues}:
We observed several cases where the LLMs gave clues using fictional words from pop culture. For example, during a game played on the board shown in Figure \ref{fig:Board_seed_1}, the o1-preview codemaster provided the clue \textit{(Hogwarts, 3)} on turn 1. The corresponding o1-preview guesser then correctly selected \textit{SCHOOL}, \textit{SPELL} and \textit{LION} from the board. As some readers may already be aware, Hogwarts is not a ``real'' word and is instead taken from the popular Harry Potter book series. As such, it is unlikely that this clue would be able to be interpreted by most traditional natural language processing techniques that rely on word embedding to see statistical associations between words. Indeed, none of the word-vector agents tested include this word in their provided corpus. Nonetheless, this clue strongly connects the three words selected by the guesser and highlights the LLMs ability to draw on cultural references as inspiration. 

\textbf{Word-vector clues rely on statistical similarity, while LLM clues use context}:
One commonly observed issue with word-vector approaches is that they often give clues which seem nonsensical to human guessers. A clue such as \textit{(Perry, 2)} that is intended for hinting at the words \textit{Mexico} and \textit{Berry}, can be easily interpreted by an identical word-vector guesser but which is largely uninterpretable for most human players. 
Word-vector agents rely on statistical similarity to generate clues, treating word associations as points on a sliding scale of relatedness. For example, the word2vec model suggested \textit{(Deer, 3)} as a clue for \textit{BUCK}, \textit{BEAR} and \textit{ROBIN}. While \textit{BUCK} is semantically cohesive with \textit{Deer}, the words \textit{BEAR} and \textit{ROBIN} are included because they are statistically closer to \textit{Deer} than the other unrelated words on the board. This process makes sense mathematically, but diverges significantly from how humans often generate and interpret clues for Codenames \cite{10.1145/3677081}.

Research in cognitive semantics suggests that humans tend to group words into cohesive categories rather than relying on statistical co-occurrence. Prototype theory highlights how humans classify concepts based on shared, salient features; and semantic priming studies demonstrate that categorical or meaningful connections are processed more quickly than weaker, statistically driven ones \cite{neelySemanticPrimingEffects1990,maclauryPrototypesRevisited1991}. Context dependence also plays a critical role in human reasoning, as associations are often shaped by situational and cultural factors rather than abstract statistical relationships \cite{barsalouContextindependentContextdependentInformation1982}.
In Codenames, human clue-givers typically prioritise clear, contextual, or semantically meaningful relationships between words. For instance, they might use \textit{Animal} or \textit{Nature} as clues to connect \textit{BUCK} and \textit{BEAR}, because these words all belong to a shared category (hypernyms). However, humans do not generate associations based on abstract statistical co-occurrence because they lack direct access to this information. Instead, their reasoning relies on context, shared knowledge, and semantic relationships that are intuitively interpretable.

This disconnect between word-vector approaches and human-generated clues may stem from differences in how humans perceive associations. Humans often expect clues to convey a reason for the connection (whether categorical, antonymic, or contextual) while word embeddings focus solely on relative semantic distances in vector space. This raises important questions for future work, such as how this mismatch in reasoning affects a user's gameplay experience, or if LLM codemasters can generate more semantically cohesive clues that improve its compatibility with human teammates?

\subsubsection{Models giving invalid responses}
One of the risks with using LLMs is their ability to give invalid responses when asked to produce clues or guesses. While this happened fairly rarely, with most invalid clues or guesses being correctly fixed with follow up prompts, we did observe some factors that appeared to increase the likelihood of invalid responses.

\textbf{Codemasters giving invalid responses}:
Invalid responses from LLM codemasters often came in the form of clue words that contained, or were contained within, another word on the board. For example, trying to give the clue \textit{Bonfire} or \textit{Campfire} when \textit{FIRE} is a word on the board would be considered an invalid response. Due to the way that our invalid clue checker was implemented, essentially checking for substring matches between the provided clue and all remaining words on the board, there is a risk for some potentially valid clues to be incorrectly flagged as invalid. For example, the clue \textit{Education} was flagged as invalid because it contained the board word \textit{CAT}, even though most human players would argue that the word education is not derived from the word cat and thus doesn't violate the game's official rules. One way to address this issue in the future would be to employ a more advanced checker, perhaps utilising a database of compound words that derive from the given clue, although this may not completely fix the problem.

\textbf{Guessers giving invalid responses}:
On the guesser's side, the Llama model would very occasionally fall into a loop of making repeatedly invalid guesses by returning a response in an invalid format. For example, when provided the clue \textit{(Transparent, 2)} Llama first tried to select the word \textit{ICE}, but this was not one of the words present on the board. When prompted to select another word, Llama repeatedly stated ``ICE is not an option, but SNOW is somewhat transparent'' instead of responding with just the individual word being selected. After being asked to guess again, Llama would rephrase the same sentence, ignoring the required single word response criteria. We suspect that because the agent sees its previous responses within its own context window, that it is inadvertently fine-tuning itself to give future invalid responses. While many of the models would eventually select one of the valid words after many repeated prompts, this is obviously not a desirable behaviour and resulted in LLMs exceeding the 10 invalid guesses limit on several occasions. 

The o1-mini guesser model would also sometimes give \textit{yes} or \textit{no} as guesses when asked to select a word on the board. This likely occurred because guesser models are also asked after each correct guess if they would like to continue guessing or stop for now, after which they need to respond with \textit{yes} or \textit{no}. If the \textit{yes} response is given, then the remaining words are presented to the model again to make a further guess. The o1-mini agent would sometimes respond with an additional yes or no after already replying to this question, possibly not realising that it was now required to make another guess. 


\subsubsection{Idiosyncrasies of model behaviour}
During our experiments, we observed several strange decisions that were repeatedly made by either the codemaster or guesser LLMs that negatively impacted their overall performance. These idiosyncratic behaviours appear to indicate some level of misunderstanding with the rules of the game by the LLMs, which hinders their strategic and epistemic reasoning capabilities.

\textbf{Over emphasis on one word}:
Firstly, many codemaster agents would overemphasise their clue's connection to one word on the board, neglecting strong connections with the other words they intended to hint at. For example, the GPT-4o model once gave the clue \textit{(Picnic, 2)} with the remaining team words to identify being \textit{PLATE}, \textit{TRUNK}, and \textit{BEACH}. While the word has strong associations with \textit{PLATE}, which was correctly guessed, connections to the other words seem tenuous at best. This clue was also given when the incorrect word \textit{JAM} was on the board, which was picked by the guesser after \textit{PLATE}. This single word prioritisation by codemasters was a repeatedly observed behaviour across all LLMs.

\textbf{Clues closely related to the assassin}:
Another strange behaviour from codemasters was vastly underestimating a given clue's potential association to the assassin word. Agents would sometimes inaccurately consider the potential relationship of their clue to other non-target words, which is especially important for the assassin word. For example, the Llama codemaster gave the clue \textit{(Spark, 2)} when \textit{BOLT} was the assassin word on the board. This issue, combined with the tendency to neglect connecting their clue to words beyond the first, would lead guesser models to often correctly guess the first word but then incorrectly pick the assassin word afterwards. 

\textbf{Continuing to guess beyond the number given by the codemaster}:
We also observed guesser models continuing to guess even after they had selected all the words related to a given clue, which is represented by the Stop Late percentage in Table \ref{results_table1}. The o1-preview and o1-mini models had the highest likelihood of this, with Stop Late percentages of 14.2\% and 16\% respectively. Humans playing Codenames would typically only go over the clue number provided by the codemaster if they can now see a connection that they previously missed. It is rare to see human players use this extra guess to randomly select a word, on the pure chance that it will be correct, outside of niche situations towards the end of the game. 

Most of the instances where the o1-preview and o1-mini models decided to continue guessing were because the guesser agent continued to find additional connections to the given clue beyond the number suggested by the codemaster. For example, in one game the clue \textit{(Wizard, 1)} was given by the codemaster, most likely to be associated with the remaining word \textit{STAFF}. This word was then correctly selected by the guesser agent, who then decided to keep guessing and selected the word \textit{HOOD}. While this may seem like a sensible guess, \textit{HOOD} was actually one of the opposing teams words. In this situation, the guesser agent may have been confused about the rules of the game, thinking that it was just looking for words connected to the given clue regardless of the provided number. This misunderstanding led the agent to continue identifying word connections and giving guesses, regardless of whether it had already reached the number provided by the codemaster.


\section{Future work}

We see many opportunities for future research that utilises Codenames as an experimental setting for evaluating the capabilities of LLMs.

To assess an LLM's ability to adapt to the opposing team's behaviour, researchers could prompt the guesser and codemaster agents to utilise information within the other team's responses as part of its own reasoning. For example, a guesser on the red team could analyse the blue team's prior clues to help determine which words are most likely to be blue. This could provide an advantage to a guesser agent, as it reduces the sample space of words they are considering. On the codemaster's side, an advanced codemaster might avoid giving clues that could be related to the opposing team's words, in order to reduce the risk of their guesser accidentally selecting them (i.e., selecting a neutral word is less bad than selecting a word for the opposing team). Choosing to leave certain target words until later in the game can also be advantageous. For example, if the red team has the word \textit{INDIA} on the board while blue team has the words \textit{CHINA}, \textit{NAPAL} and \textit{JAPAN}, then by choosing not to give a clue for \textit{INDIA} until much later in the game the red codemaster limits the potential of the blue codemaster to give easy clues such as \textit{Country} or \textit{Asia}. It is also possible that the blue guesser may accidentally select \textit{INDIA} when attempting to identify the other words. Focusing on an LLMs ability to perform such reasoning may reveal whether they can effectively take other agent's behaviour and circumstances into account when making strategic decisions.

Codenames also provides a rich and controlled setting to investigate the linguistic reasoning capabilities of LLMs. We see potential for many possible extensions of our work that delves deeper into the specific linguistic tendencies of LLMs. Words can be linguistically connected in many ways, such as synonyms (happy, joyful), antonyms (hot, cold), hyponyms (dog to animal) and hypernyms (vehicle to car). Investigating the ways in which LLM codemasters tend to relate words, and also how successfully guesser agents can connect words using the provided clue, may lead to better ways of communicating with LLM agents in other settings.

Additional experiments could provide further comparisons on how different techniques (including both LLM and word-vector approaches) can be most effectively combined. For example, while the current word-vector agents cannot reliably act as guesser when paired with an LLM, due to their inability to interpret words outside of their provided corpus, they can still operate as Codemaster alongside an LLM guesser. Such a comparison could help to indicate if any of the word-vector approaches provide clues that are more human interpretable, as LLMs have previously been shown to work well at collaborating alongside human players \cite{sidjiHumanAICollaborationCooperative2024}.

The type of words on the board is another factor that can be controlled and varied for different agents. Custom word-pools can be developed to test an LLM's ability to make linguistic connections. For example, we could create a board where all words are very closely related to the assassin word, which is intended to evaluate an agent's ability to distinguish between heavily related words. Boards with words that are fictional or related to a specific topic could also be used to test an agent's ability to relate specific cultural references. Broadening beyond words, Codenames also has a picture-based version, where 25 picture cards replace the 25 words on the board. This version could be used to test the multimodal reasoning of LLM agents equipped with image understanding capabilities. To further extend this, we could also experiment with multimodal agents giving clues in the form of generated images. 

In addition to Codenames, many other language-based boardgames could also be explored in future studies to help verify if our LLM performance and behavioural findings are specific only to codenames or are more general. Some example games that could be investigated further include ``So Clover'', `Just One'', ``Letter Jam'', ``Decrypto'', ``Medium'' and ``Master Word''. If LLM agents can be developed with a high level of competence in these language-based games, we may also be able to extract explainable strategies by studying their gameplay. This approach is already being taken in other AI dominated games such as Go, Chess and Dota 2 \cite{shinHumanLearningArtificial2021}.

\section{Conclusion}

In this paper we have explored the potential of the game of Codenames as a suitable benchmark for assessing the language reasoning capabilities of multiple LLM agents, as well as their strategic tendencies within the game. We did this by having different LLMs play both the single team and two team version of Codenames, alongside a variety of different teammates and opponents. We found that each LLM exhibits a unique emergent style of play and that they do not necessarily perform best when paired together. In fact, each LLM often performs better in a specific role (either as codemaster or guesser). We found that a cautious playstyle often resulted in high performance for the single team version, with the exception of the new o1-preview model which was able to achieve superior performance over all other LLMs despite exhibiting a more risky strategy. For the two teams version the opposite appeared to be the case, where playing risky often led to a higher average win-rate against a more cautious opponent. We also discuss why LLM agents are more suitable for human play than previous word-vector approaches, observed idiosyncrasies in model behaviour, and the rich potential for future work in LLM research utilising Codenames. 

\bibliographystyle{IEEEtran}
\bibliography{bib}

%

\begin{IEEEbiography}[{\includegraphics[width=1in,height=1.25in,clip,keepaspectratio]{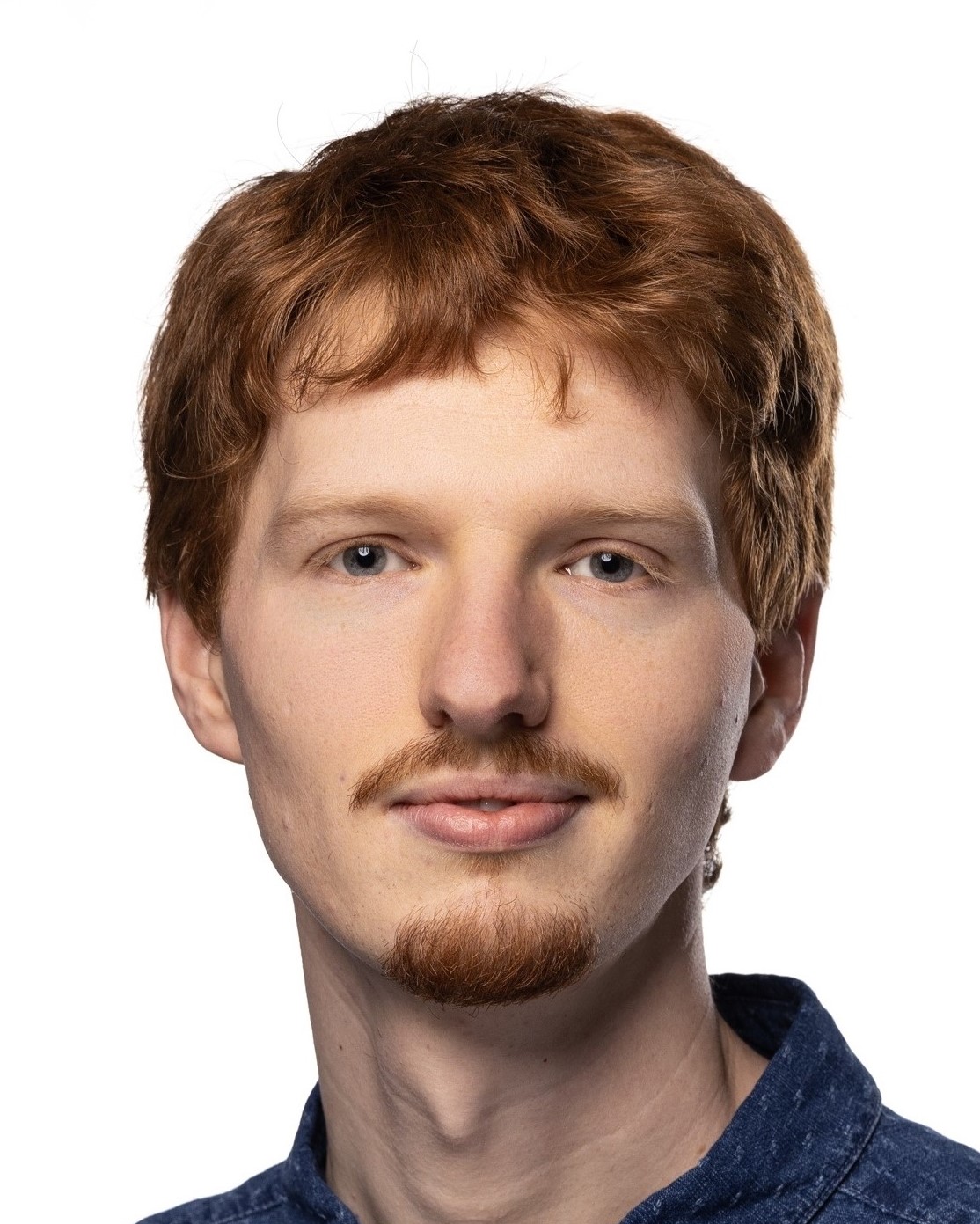}}]{Matthew Stephenson}
Dr. Matthew Stephenson is a Lecturer with the College of Science and Engineering at Flinders University in South Australia. His research focusses on applying Artificial Intelligence, Machine Learning and Data Science techniques to games. This includes designing AI to play, create and analyse games; as well as utilising games as a testbed for developing AI-based solutions to real-world problems.
\end{IEEEbiography}

\begin{IEEEbiography}[{\includegraphics[width=1in,height=1in,clip]{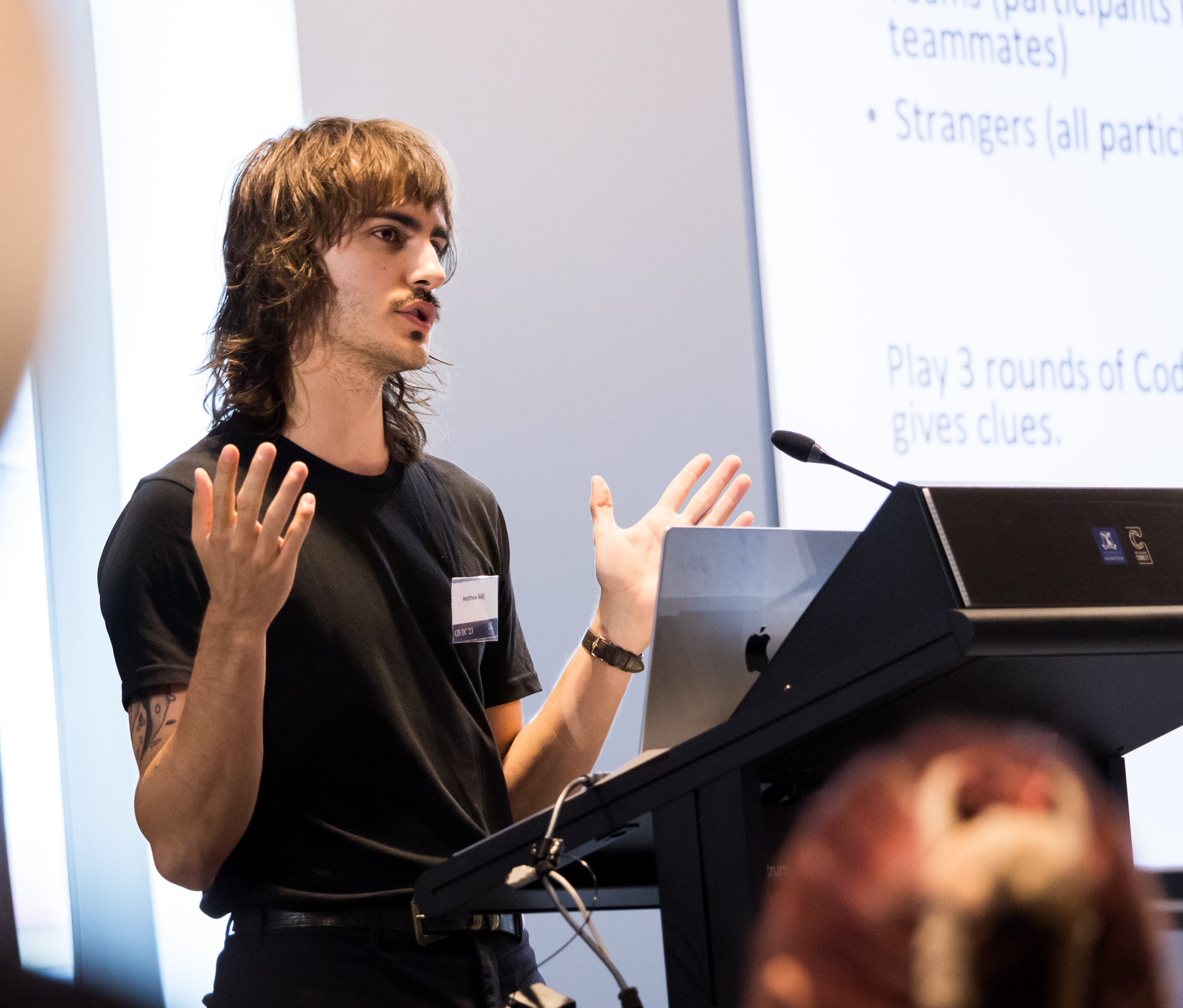}}]{Matthew Sidji}
Matthew Sidji is a final year PhD candidate with the School of Computing and Information Systems at The University of Melbourne in Victoria. His research focusses on AI's affect on human teaming and cognition in cooperative games. His work involves investigating human play practices and developing cooperative AI agents for games.
\end{IEEEbiography}


\begin{IEEEbiography}[{\includegraphics[width=1in,height=1in,clip]{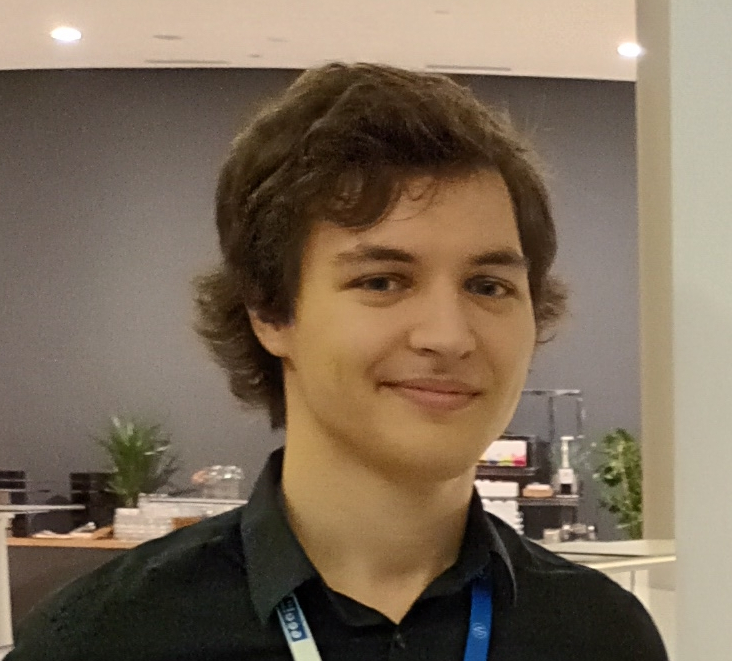}}]{Benoît Ronval}
Benoît Ronval is a PhD student with the Institute of Information and Communication Technologies, Electronics and Applied Mathematics (ICTEAM) at UCLouvain in Belgium. His research centers on generating synthetic tabular data and its integration with LLMs. He specializes in designing generative models and investigating the applications of synthetic data in machine learning.
\end{IEEEbiography}




\end{document}